\pdfoutput=1

\documentclass[11pt]{article}

\usepackage{ACL2023}

\usepackage{times}
\usepackage{latexsym}
\usepackage{graphicx} 
\usepackage{amsmath, bm}
\usepackage[T1]{fontenc}

\usepackage[utf8]{inputenc}

\usepackage{microtype}

\usepackage{inconsolata}

\newcommand{\hide}[1]{} 
\newcommand{\vpara}[1]{\vspace{0.05in}\noindent \textbf{#1 }}

\newcommand{\model}{FC-KBQA}
\newcommand{\smodel}{FC-KBQA }

\title{\model: A Fine-to-Coarse Composition Framework for Knowledge Base Question Answering}

\author{Lingxi Zhang\textsuperscript{1}, Jing Zhang\textsuperscript{1}\thanks{       $\text{ }$ \scriptsize Corresponding author.\normalsize}, Yanling Wang\textsuperscript{1}, Shulin Cao\textsuperscript{2}, Xinmei Huang\textsuperscript{1}, \\\textbf{Cuiping Li\textsuperscript{1},} \textbf{Hong Chen\textsuperscript{1},} \textbf{Juanzi Li\textsuperscript{2}} \\
  \textsuperscript{1}School of Information, Renmin University of China, Beijing, China\\
\textsuperscript{2}Department of Computer Science and Technology, Tsinghua University, Beijing, China\\
  \{zhanglingxi, zhang-jing, wangyanling,huangxinmei, licuiping, chong\}@ruc.edu.cn\\
  \{caosl19\}@mails.tsinghua.edu.cn,
  \{lijuanzi\}@tsinghua.edu.cn
  }

\begin{document}
\maketitle
\begin{abstract}
The generalization problem on KBQA has drawn considerable attention. Existing research suffers from the generalization issue brought by the entanglement in the coarse-grained modeling of the logical expression, or inexecutability issues due to the fine-grained modeling of disconnected classes and relations in real KBs. We propose a Fine-to-Coarse Composition framework for KBQA (\model) to both ensure the generalization ability and executability of the logical expression. The main idea of \smodel is to extract relevant fine-grained knowledge components from KB and reformulate them into middle-grained knowledge pairs for generating the final logical expressions.
\smodel derives new state-of-the-art performance on GrailQA and WebQSP, and runs 4 times faster than the baseline. 
Our code is now available at GitHub \url{https://github.com/RUCKBReasoning/FC-KBQA}.

\end{abstract}

\section{Introduction}
\label{sec:introduction}
\begin{figure}[t]
\centering
\includegraphics[scale=0.33]{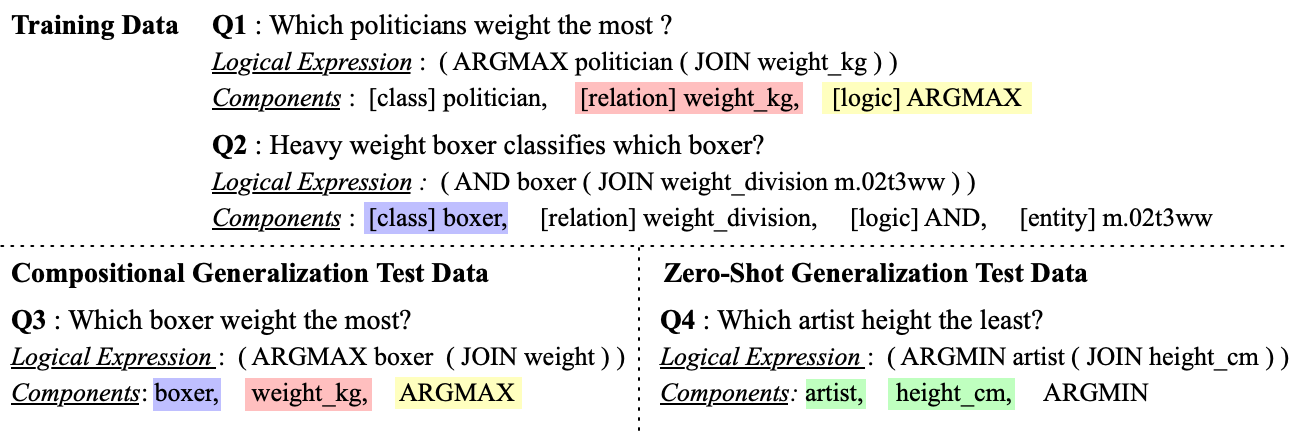}
\caption{Illustration of generalization tasks in KBQA. Each question is paired with a logical expression that consists of different components. Components involved in the training data are colored in non-green color, while unseen components are colored in green.}     
\label{fig:dataset}

\end{figure}

Question answering over knowledge bases (KBQA) aims to provide a user-friendly way to access large-scale knowledge bases (KBs) by natural language questions. Existing KBQA methods~\cite{ZHANG20231} can be roughly categorized into retrieval-based and semantic-parsing (SP) based methods. The former ~\cite{feng-etal-2021-pretraining-numerical,he2021improving,zhang2022subgraph} directly scores the relevance between the question and answer candidates, thus
it is difficult to resolve the complex questions. On the contrary, some KBQA approaches, such as \cite{das2021case,kapanipathi2021leveraging,qiu2020stepwise,sun2020sparqa}, are based on semantic parsing (denoted as SP-based), which can address complex questions and achieve promising results on i.i.d. datasets. SP-based methods first translate the questions into logical expressions such as SPARQL and then execute them against KB to yield answers. As illustrated in Figure~\ref{fig:dataset}, a logical expression consists of multiple components such as classes and relations. Most existing SP-based approaches fail with logical expressions that contain unseen compositions of components (called compositional generalization) or unseen components (called zero-shot generalization).

\begin{figure}[t]
\centering
\includegraphics[scale=0.38]{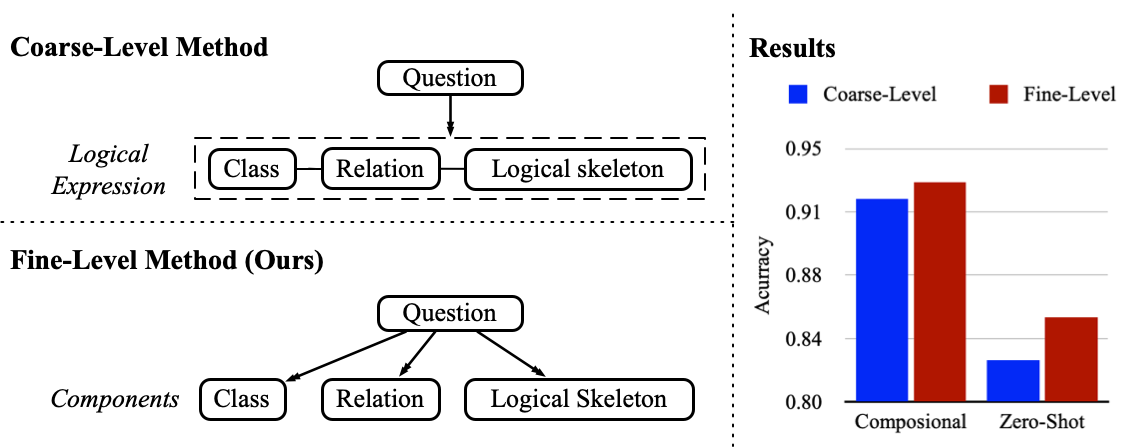}
\caption{Results of the pilot study. The coarse-grained method directly matches the question with the logical expression (i.e., the composition of components), while the fine-grained method matches the question with each component candidate and then composes them to derive the logical expression. The exact match accuracy of logical expressions on compositional generalization test data and zero-shot generalization test data is shown on the right of the figure.}     
\label{fig:pre_experiment}

\end{figure}

To address the above problem, GrailQA-Rank~\cite{gu2021beyond} proposes a BERT-based rank model to match the given question with each logical expression candidate, which leverages the generalization abilities of the pre-trained language models. 
On top of that, RNG-KBQA~\cite{ye2022rng} further uses a pre-trained generation model, which takes top-5 ranked logical expressions as the additional input beyond the question to generate the target logical expression.  
Behind these mainstream models, a logical expression is viewed as an inseparable unit during modeling. Actually, logical expressions are coarse-grained because they can be decomposed into relatively fine-grained components including relations, classes, entities, and logical skeletons (See examples in Figure~\ref{fig:KB}). 
Such coarse-grained modeling entangles representations of fine-grained components, thereby overfitting the seen compositions during the training process, which weakens the model's compositional generalization ability. 
Meanwhile, even though pre-trained language models can deal with zero-shot components to some extent, compositional overfit reduces their ability to identify individual unseen components with zero-shot generalization. 

To demonstrate the above idea, we perform a pilot study (Cf. the detailed settings in Section~\ref{sec:pre-experiment}) with two preliminary experiments: one calculates the similarity score between a question and each coarse-grained logical expression to obtain the most relevant one, and the other searches the most relevant fine-grained components to form the final logical expression of a question. We observe that \textbf{the fine-grained modeling derives more accurate logical expressions on both the compositional task and zero-shot task (Cf. Figure~\ref{fig:pre_experiment})}. It could be explained that fine-grained modeling focuses exclusively on each component, avoiding overfitting of seen compositions in the training data. 
Although some studies attempt to leverage fine-grained components, they only consider partial fine-grained components such as relations, classes, and entities ~\cite{chen-etal-2021-retrack}, or suffer from inexecutability due to disconnected fine-grained components in real KBs~\cite{shu2022tiara}.

Thus, to both ensure the generalization ability and executability of logical expressions, we propose a \underline{F}ine-to-\underline{C}oarse composition framework for KBQA (\model), which contains three sub-modules. The overview of our model is shown in Figure~\ref{fig:over-view}.
The first module is fine-grained component detection, which detects all kinds of fine-grained component candidates from Freebase by their semantic similarities with the question. Such component detection guarantees the generalization ability in both compositional and zero-shot tasks. The second module is the middle-grained component constraint, which efficiently prunes and composes the fine-grained component candidates by ensuring the components' connectivity in the KB. 
The final module is the coarse-grained component composition, which employs a seq-to-seq generation model to generate the executable coarse-grained logical expression. In addition to encode the fine-grained components, the middle-grained components are also encoded to enhance the model's reasoning capacity, so as to improve the executability of the generated logical expression. In contrast to previous work~\cite{cao-etal-2022-program,chen-etal-2021-retrack,shu2022tiara} that only uses the knowledge constraints to guide the decoding process, we emphasize injecting them into the encoding process, because the encoder which learns bidirectional context could better suit natural language understanding~\cite{du2022glm}.

We conduct extensive experiments on widely used GrailQA, WebQSP, and CWQ datasets. GrailQA~\cite{gu2021beyond} is a KBQA benchmark focusing on generalization problems. \smodel derives new state-of-the-art performance on GrailQA-Dev (+7.6\% F1 gain and +7.0\% EM gain respectively). Meanwhile, \smodel also obtains good performance on WebQSP and CWQ. Moreover, \smodel runs 4 times faster than the state-of-the-art baseline RNG-KBQA. The ablation studies demonstrate the effect of our middle-grained encoding strategy.

\vpara{Contributions.}
(1) We conduct a pilot study to reveal an intriguing phenomenon --- a fine-grained understanding of the logical expression helps enhance the generalization ability of SP-based KBQA methods, which is rarely discussed before.
(2) We propose a fine-to-coarse composition framework \smodel to address the generalization problem, which takes advantage of the idea of fine-grained modeling.
(3) We devise a middle-grained component constraint that is injected into both the encoder and the decoder to guide the seq-to-seq model in producing executable logical expressions.
(4) \smodel not only maintains efficiency but also achieves significant improvement on GrailQA.

\section{Related Work}
\vpara{Coarse-Grained SP-based Methods.}
Many efforts are paid to solve generalization problems on SP-based KBQA. Some approaches, such as~\cite{lan2020query,gu2021beyond}, use a rank-based model that takes advantage of a coarse-level match between the question and the logical expressions or query graphs. They first enumerate numerous query graph candidates based on KBs and then they rank them according to how relevant they are to the question. Another line of approaches, in addition to the rank-based ones, makes use of a generation model. KQAPro~\cite{cao2022kqa} leverages BART to directly convert questions into logical expressions. Additionally, RNG-KBQA~\cite{ye2022rng} further injects top-k ranked logical expressions as an additional input to the question. CBR-KBQA~\cite{das2021case} injects analogous questions and their corresponding logical expressions from the training data to increase the generalization. All of the aforementioned methods are pure coarse-level frameworks that treat each coarse-grained logical expression as a separate unit. 

\vpara{Fine-Grained SP-based Methods.}  
Many researchers have been motivated to address the generalization issue by the notion of utilizing decomposed components, such as class, relation, and logical skeleton.
Some approaches~\cite{wang-etal-2020-rat,zhao-etal-2022-bridging,li2022resdsql} retrieve the relevant schema item such as relation and column as additional fine-grained input information, while another line of approaches~\cite{dong-lapata-2018-coarse} extracts the skeleton of logical expression as the decoder guide. Such methods primarily concentrate on the grammar of logical expression and often ignore the knowledge constraint, which is essential in large-scale KB. They usually focus on KBs or DBs that contain a small number of relations where a logical expression can be easy to be executable. Program Transfer~\cite{cao-etal-2022-program}, ReTrack~\cite{chen-etal-2021-retrack}, and TIARA~\cite{shu2022tiara} simply apply KB constraints to control the generation of the decoding process. As opposed to them, we make use of middle-grained KB constraints during both the encoding and the decoding processes to help the model better adapt to KB and ensure executability.

\section{Problem Definition}
\label{sec:problem}
\begin{figure}
\centering
\includegraphics[scale=0.58]{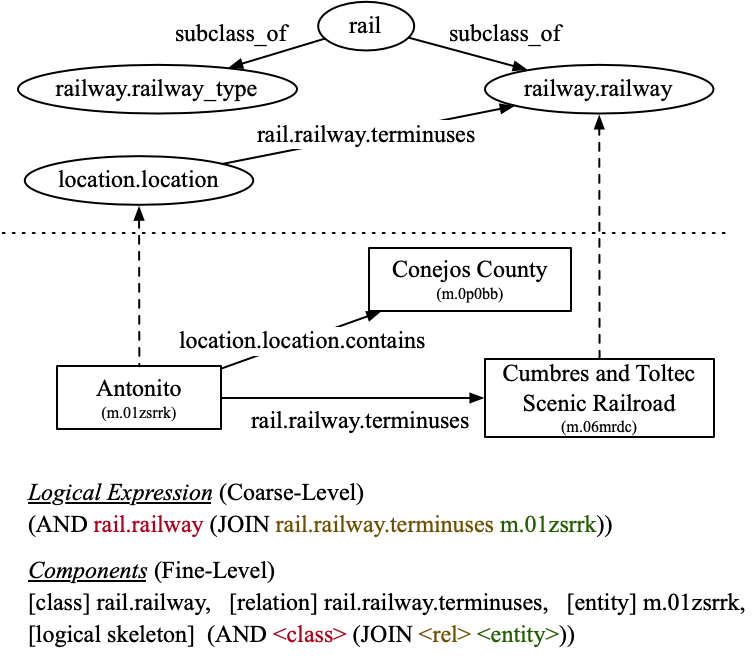}
\caption{Illustration of a KB subgraph and an executable logical expression, where the ovals denote the entities, the rectangles denote the classes, the solid lines denote the relations, and the dashed lines connect the entities and their classes.
The upper part of the subgraph illustrates examples of ontology triplets, while the bottom illustrates relational facts.
}     
\label{fig:KB}
\end{figure}

\vpara{Knowledge Base (KB).}
A KB is comprised by ontology $\{(C \times R \times C)\}$ and  relational facts $\{(E \times R \times (E\cup C))\}$, where $R,C,$ and $E$ denote relation set, class set, and entity set respectively.
Notably, we consider literal as a special type of entity.
Specifically, an ontology triple $(c_d, r, c_r)$ consists of a relation $r \in R$, a domain class $c_d$ which denotes the class of the subject entities, and a range class $c_r$ which denotes the class of the object entities.
Each class has multiple entities, thus an ontology triplet can be instantiated as several relational facts. For example, both $(e_1,r,e_2)$ and $(e_3,r,e_4)$ correspond to $(c_d, r, c_r)$, where $e_1, e_3 \in c_d$ and $e_2, e_4 \in c_r$.
Figure~\ref{fig:KB} illustrates a KB subgraph.

\vpara{SP-based KBQA.}
Given a natural question $q$, KBQA models aim to find a set of entities denoted by $A 
\subseteq E$ from KB as the answers to $q$. Instead of directly predicting $A$, SP-based KBQA models translate $q$ to an executable logical expression denoted by $s$ such as SPARQL, lambda-DCS~\cite{liang2013learning}, query graph~\cite{lan2020query}, and s-expression~\cite{gu2021beyond}. 

We select s-expression as our used logical expression since it could provide a good trade-off on compactness, compositionality, and readability~\cite{gu2021beyond}.
The \textbf{logical skeleton} of an s-expression can be derived by removing all the relations, classes, and entities in the expression and only keeping function operators and parentheses. 
Specifically, we replace relations, classes, entities, literals with special tokens ``<rel>'', ``<class>'', ``<entity>'', ``<literal>'' respectively. 
Figure~\ref{fig:KB} shows an executable logical expression on the KB and its corresponding logical skeleton. We unitedly name the relations, classes, entities, and logical skeleton in an s-expression as the \textbf{fine-grained component}, while the complete s-expression is the \textbf{coarse-grained logical expression}.

\section{Approach}
\begin{figure*}[t]
\centering
\includegraphics[scale=0.63]{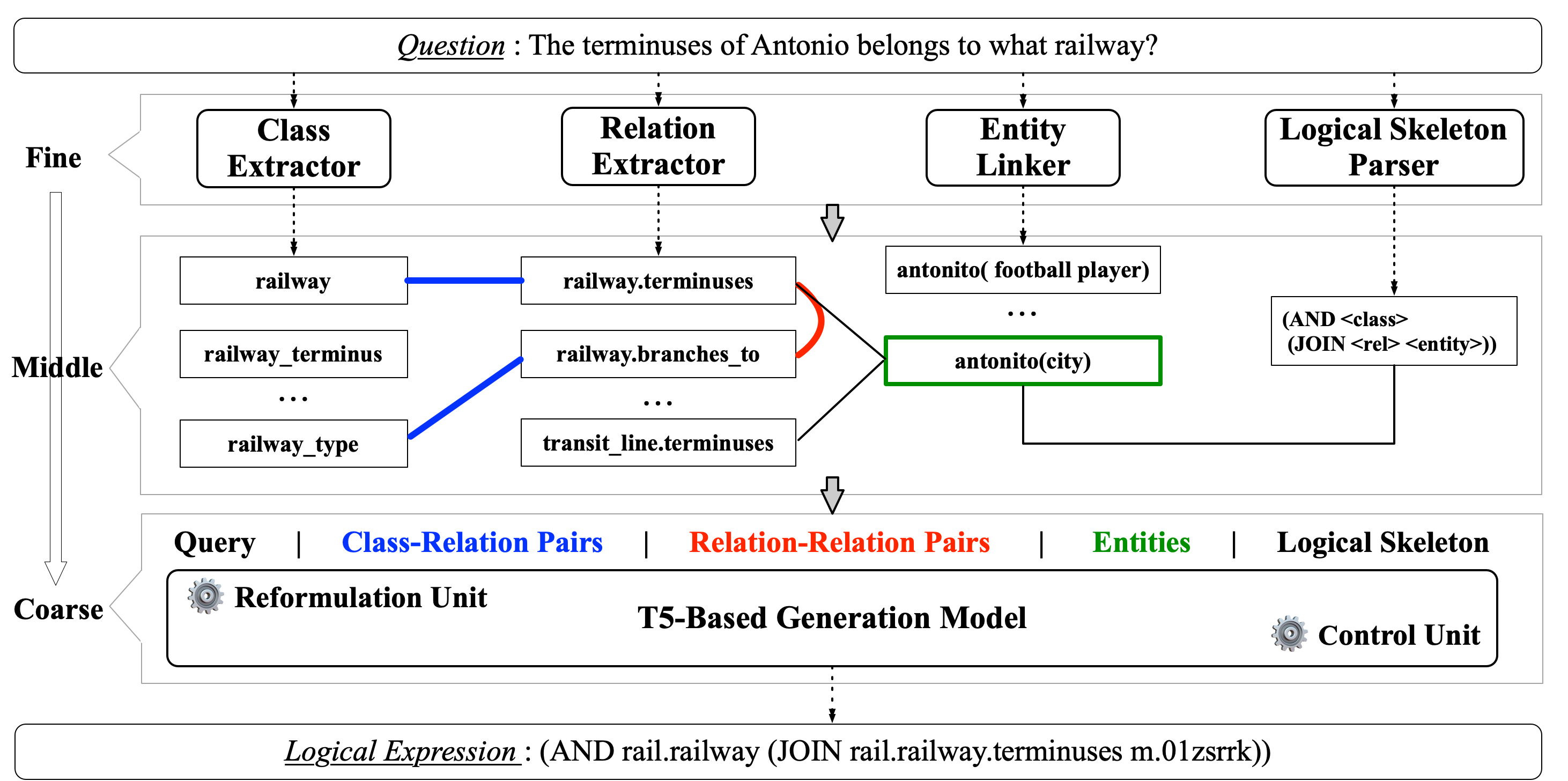}
\caption{Overview of \model. In the step of fine-grained component detection, we perform class extraction, relation extraction, entity linking, and logical skeleton parsing to obtain the most relevant components of the question. Then we utilize the KB-based constraint to obtain middle-grained component pairs that are connected in the KB. Finally, a T5-based seq-to-seq model encodes the reformulated fine-grained and middle-grained candidates (reformulation unit), and employs a controllable decoder with dynamic vocabulary (control unit) to generate the executable target logical expression.}
\label{fig:over-view}
\end{figure*}

\subsection{Pilot Study}
\label{sec:pre-experiment}
As analyzed in Section~\ref{sec:introduction}, considering the logical expression as a unit will lead to entangled representations of fine-grained components and thus weakens generalization ability. Here we study the necessity of fine-grained modeling by testing how coarse-grained and fine-grained matching methods perform when selecting a question's logical expression from the corresponding candidate pool.

\vpara{Dataset.}
To simplify the experiment, we extract a toy dataset that only involves 1-hop logical expressions from GrailQA.
Then, for the relation $r$ and the class $c$ in such logical expressions, we study the compositional generalization where the composition $(r,c)$ is unseen or zero-shot generalization where the individual $r$ or $c$ is unseen in the training data.
For each question with its ground-truth logical expression, we select 100 logical expressions that share the same domain as the ground truth as the coarse-grained expression candidates. For fair comparison, we separate all of the relations, classes, and logical skeletons from the coarse-grained candidates as the fine-grained component candidates.

\vpara{Methods.}
We aim to find the target logical expression of a given question by a ranking model trained with a contrastive loss~\cite{chen2020simple}, which is also used by RNG-KBQA~\cite{ye2022rng}.
The coarse-grained method concatenates a question and a candidate logical expression to feed into BERT, then the output embedding of [CLS] is fed into a linear layer to compute the similarity score.
The fine-grained method follows the above pipeline, but the input is the concatenation of a question and a fine-grained candidate component, then scores each logical expression candidate by summing up the normalized question-component similarity scores. 
For both methods, we compute accuracy by evaluating whether the ground-truth logical expression owns the highest score in the candidate pool.

\vpara{Observation --- Fine-grained modeling can better solve the generalization problems on KBQA.} The matching accuracy is reported in Figure~\ref{fig:pre_experiment}. The fine-grained method outperforms the coarse-grained method in both composition generalization and zero-shot generalization tasks. 
A possible explanation is the fine-grained matching focuses solely on each component and is simple to learn, which better capture the semantic information of each component and also well adaptable to express the various compositions of components. The coarse-grained matching, on the other hand,  attempts to describe all of the components as a whole composition, limiting the ability to express unseen compositions and components.
Inspired by this, we propose \smodel in the next section.

\subsection{Model Overview}
We propose a fine-to-coarse composition framework \smodel bridged by a middle-grained KB constraint. Figure~\ref{fig:over-view} illustrates the overall framework, which contains three parts:

\vpara{Fine-grained Component Detection.} Given a question, we extract relation candidates and class candidates from the whole KB based on semantic similarity. Simultaneously, we adopt an entity linker to detect mentioned entities and use a seq-to-seq model to generate logical skeletons. 

\vpara{Middle-grained Component Constraint.}
Based on the detected components, we devise an efficient way to check the connectivity of component pairs on the KB, including class-relation pairs, relation-relation pairs, and relation-entity pairs. We only keep the executable component pairs to guarantee the executability of final logical expression. 

\vpara{Coarse-grained Component Composition.} 
Finally, a seq-to-seq model takes the concatenation of the question and the reformulated components as input to generate the logical expression. In particular, the middel-grained components are injected into both the encoder and the decoder to ensure the executability of the final logical expressions.


\subsection{Fine-grained Component Detection}
\label{sec:fine_grained}
\vpara{Relation and Class Extraction.}
Taking the relation extractor as the example, given a question $q$, we aim to extract relations in $q$. First, we apply BM25~\cite{robertson2009probabilistic} to recall the relation candidates from the KB based on the surface overlaps between relations' names and $q$. 
Then we apply BERT~\cite{devlin2018bert} as the cross-encoder to measure the semantic similarity between $q$ and each relation candidate $r$. 
We describe $r$ using the relation domain, the relation name, and the relation range and let the BERT input be ``[CLS] q [D] domain(r) [N] name(r) [R] range(r) [SEP]'', where [CLS], [SEP], [D], [N], and [R] are the special tokens. 
To better distinguish the spurious relations, we sample the relations that share the same domain as the ground-truth relation as the negatives for training.
The trained model is used to retrieve the set of top-$k$ relations, denoted by $R_q$.

The class extractor works in the same way as the relation extractor. We represent the class using its name and domain, and use other classes in the same domain as negatives. $C_q$ represents the set of the top-$k$ relevant classes.

\vpara{Entity Linking.}
A common paradigm of finding topic entities in KBQA methods is to first leverage a NER tool~\cite{finkel2005incorporating} to detect mentions and then apply an entity disambiguation model to link them to entities in KB. However, some noun-phrase mentions such as ``rich media'' are hard to be detected by the NER tool, and some ambiguous entities could not be distinguished by the pure entity names.  To address both issues, we equip the NER tool\footnote{We follow GrailQA which utilizes an open BERT-NER tool on GitHub (https://github.com/kamalkraj/BERT-NER).} with a trie tree-based mention detection method and propose a relation-aware pruning method to filter the mentions.

Specifically, we build a trie tree~\cite{fredkin1960trie} with the surface names of all entities in the KB. Then we can search noun phrase mentions in the question efficiently and link them to the KB by BLINK~\cite{wu2019zero} to obtain the corresponding entities $E_q$. After that, we propose a relation awared pruning strategy to prune $E_q$ by removing the entities that could not link to any relations in $R_q$. Finally, following GrailQA~\cite{gu2021beyond}, we choose the entity with the highest popularity. 
We define regular expressions to extract literals such as digits and years appearing in $q$.

\vpara{Logical Skeleton Parsing.}
Logical skeleton parsing aims to transform a given question $q$ into a logical skeleton $l$. Because the logical skeleton is domain-independent, the parsing process could be generalized across domains. We adopt T5~\cite{raffel2020exploring}, a state-of-the-art generation model to parse logical skeletons. Since many entity names contain tokens such as ``and'' and
``of'' that may cause the logical skeleton to be incorrectly determined, we mask each mention $m \in M_q$ with the special token ``<entity0>'', ``<entity1>'', ..., in order of appearance. For example, we change ``Thomas was the designer of what ship?'' to ``<entity0> was the designer of what ship?''. We notice that a common error is parsing out logical skeleton with wrong relation numbers, for example ``<rel>'' instead of ``<rel><rel>''. Instead of increasing beam numbers, we manually add grammar rules, such as add ``<rel><rel>'' as the second candidate when ``<rel>'' is T5's top-1 prediction. The set of the top-2 logical skeleton candidates is denoted as $L_q$.



\subsection{Middle-grained Component Constrain}

After deriving the candidate components according to Section~\ref{sec:fine_grained},
the KB-based constraint is required to guarantee the composed logical expression is executable. 
A straightforward idea is to fill the logical skeleton with candidate relations, classes, and entities, and execute them one by one to check executability.
However, such enumeration is inefficient, since all combinations of candidate components should be considered. Therefore, we incorporate the middle-grained component pairs which are connected in KB. Such pairs can be produced efficiently to keep the model's efficiency.

The middle-grained component pairs include class-relation pairs, relation-relation pairs, and relation-entity pairs. For each class $c \in C_q$ and each relation $r \in R_q$, if $r$ is connected with the domain class $c$, we add $(c, r)$ into the class-relation pair set $P_{c-r}$. For example in Figure~\ref{fig:KB}, the class ``railway.railway'' is linked with the relation ``rail.railway.terminuses'', so the pair (railway.railway, rail.railway.terminuses) is executable and will be added into $P_{c-r}$. If the range class of $r$ is $c$, we add the pair of $c$ and the reverse relation of $r$. 
We construct executable relation-relation pair set $P_{r-r}$ by checking each relation pair $(r_1\in R_q, r_2\in R_q)$. If $r_2$'s domain class does not match $r_1$'s range class, we directly remove this pair to maintain efficiency, otherwise, we reformulate $(r_1, r_2)$ to a logical expression and execute on KB to check its connectivity. 
For each relation-entity pair $(r, e)$, we first check whether the logical skeleton candidates contain the <entity> placeholder or not. If not, we leave $P_{r-e}$ empty; otherwise we directly take the result of the relation-pruning strategy for entities in Section~\ref{sec:fine_grained}. 

\subsection{Coarse-grained Component Composition}
We apply a generation model based on T5 to compose all the above fine-grained and middle-grained component candidates and output an executable logical expression by a controlled decoder.

\vpara{Encoding Process.}
Before feeding the fine-grained and middle-grained component candidates into the generator, we sort the middle-grained candidates according to their similarity scores to the question. By doing this, the order can reveal the pattern of which pair is more likely to appear in the ground-truth logical expression. In intuition, such a pattern will help to generate more accurate logical expressions.
To accomplish this, we take the logits of the fine-grained component detection in section~\ref{sec:fine_grained} as the similarity score between the question and each class/relation component, and then calculate the similarity score between the question and a middle-grained component pair by summing the scores of contained single components. The encoding of such middle-grained component improves the generator's reasoning capacity in terms of capturing the knowledge constraints. 


We use ``;'' to separate each element (a component or a component pair). To explicitly inform the model the type of each component, we place ``[REL]'', ``[CL]'', ``[ENT]'', and ``[LF]''  before each relation, class, entity, and logical skeleton respectively. For example, we organize the input of encoder as 
``query;[CL]$c_1$[REL]$r_1$;[REL]$r_1$ [REL]$r_2$;[CL]$c_2$[REL]$r_3$;[ENT]$e_1$;[LF]$l_1$;[LF]$l_2$''.


\vpara{Decoding Process.}
The middle-grained components are also used to produce a dynamic vocabulary to constrain the decoding process.
The generated token $y_t$ is confined to the tokens involved in the dynamic vocabulary at each step $t$.
We initialize the dynamic vocabulary with the union of tokens from the detected entities, tokens from the detected classes in $P_{c-r}$, i.e., usually the answer type, and the keywords such as ``JOIN'' in logical skeleton.
Then we update the dynamic vocabulary by the relations paired with $r$ in $P_{r-r}$ if the last generated component is $r$ or by the relations paired with $c$ in $P_{c-r}$ if it is $c$.



\section{Experiment}
\begin{table*}[t]
\centering
\caption{
    Results of overall evaluation on GrailQA-LeaderBoard (\%).
	\label{tb:overall} 
}
\begin{tabular}{ccccccccc}
\hline
              & \multicolumn{2}{c}{Overall} & \multicolumn{2}{c}{I.I.D.} & \multicolumn{2}{c}{Compositional} & \multicolumn{2}{c}{Zero-Shot} \\ \cline{2-9} 
              & EM           & F1           & EM           & F1          & EM              & F1              & EM            & F1            \\ \hline
GrailQA-Rank~\cite{gu2021beyond}  & 50.6         & 58.0         & 59.9         & 67.0        & 45.5            & 53.9            & 48.6          & 55.7          \\
GrailQA-Trans~\cite{gu2021beyond} & 33.3         & 36.8         & 51.8         & 53.9        & 31.0            & 36.0            & 25.7          & 29.3          \\
ReTrack~\cite{chen-etal-2021-retrack}       & 58.1         & 65.3         & 84.4         & 87.5        & 61.5            & 70.9            & 44.6          & 52.5          \\
RNG-KBQA~\cite{ye2022rng}      & 68.8         & 74.4         & 86.2         & 89.0        & 63.8            & 71.2            & 63.0          & 69.2          \\ \hline
\model (Ours)      & \textbf{73.2}            & \textbf{78.7}            & \textbf{88.5}            & \textbf{91.2}           & \textbf{70.0}               & \textbf{76.7}              & \textbf{67.6}             & \textbf{74.0}             \\ \hline
\end{tabular}
\label{tab:overall_eval_GrailQA}
\end{table*}

\subsection{Experimental Settings}
\vpara{Dataset.}
We evaluate our method on GrailQA~\cite{gu2021beyond}, WebQSP~\cite{yih2016value}, and CWQ~\cite{talmor-berant-2018-web}, all of which are based on Freebase. GrailQA focuses on generalization problems which involved up to 4-hop logical expressions and complex operations. WebQSP is an i.i.d. benchmark that required 2-hop reasoning. Although CWQ is not designed to solve generalization problem, we can still separate out the zero-shot test set with all the unseen relations and classes, yielding 576/3519 zero-shot/all test set. 

\vpara{Evaluation Metrics.}
To measure the accuracy of logical expression, we use the well-adopted exact match (EM) which measures the exact equivalence between the query graph of the predicted and the gold logical expression. We also calculate the F1 score based on the predicted and gold answers. 

\vpara{Baselines.}
On GrailQA, we mainly compare with the published works on the leaderboard, including GrailQA-Rank~\cite{gu2021beyond}, GrailQA-Trans~\cite{gu2021beyond},  Retrack~\cite{chen-etal-2021-retrack}, RNG-KBQA~\cite{ye2022rng}. They are all SP-based models that target generalization problems in KBQA.
On WebQSP and CWQ, we compare our method with the retrieval-based models including GraphNet~\cite{pu2018graphnet},PullNet~\cite{sun2019pullnet} and NSM~\cite{He-WSDM-2021}, and the SP-based models including QGG~\cite{lan2020query}, RNG-KBQA~\cite{ye2022rng}, and PI Transfer~\cite{cao-etal-2022-program}.
We evaluate F1 for the retrieval-based models, while evaluate both F1 and EM for the SP-based methods. We compare all the baselines that have the results on the two datasets or publish the codes that can be executed.

\subsection{Overall Evaluation}
\vpara{Performance.}
In Table~\ref{tb:overall} and Table~\ref{tb:webqsp}, we evaluate the performance of \smodel on different datasets. 
For the baselines, we directly take their results reported in the original papers. To be noted, on the extracted zero-shot test set of CWQ, the results for some models remain empty because their full codes are not released.
As shown in Table~\ref{tab:overall_eval_GrailQA}, our model outperforms all the baselines, especially on the compositional and zero-shot test tasks. Compared with RNG-KBQA, the state-of-the-art published model, we have an absolute gain of 4.3\% and 4.4\% in terms of F1 score and EM respectively. We also outperform on the extracted zero-shot CWQ test set by 11.3\% in terms of F1, as for an unseen complex question, parsing out correct knowledge components and logical skeletons is much easier than directly parsing the coarse-grained logical expression correctly.
Since the fine-grained module solely focuses on each component and thus leads to a higher component accuracy, \smodel also outperforms on the i.i.d test set of WebQSP. On the original test set of CWQ, we only under-perform PI Transfer which leverages a pre-train process on a large-scale wiki data that is out scope of CWQ. 




\begin{table}\centering
\caption{\label{tb:webqsp} F1 Evaluation on WebQSP and CWQ (\%).}
\small
\begin{tabular}{lc|cc}
\hline
 & WebQSP & \multicolumn{2}{c}{CWQ} \\ \cline{2-4} &Overall    & Overall      & Zero-Shot    \\ \hline
GraphNet                                            & 66.4   & 32.8 & 22.3          \\
PullNet & 68.1   & 47.2 & -             \\
NSM    & 74.3   & 48.8 & 31.6          \\ \hline
QGG      & 74.0   & 40.4 & 28.9          \\
RNG-KBQA     & 75.6   & 42.3 & 33.3          \\ 
PI Transfer                        & 76.5   & \textbf{58.7} & -             \\ \hline
Ours                                                & \textbf{76.9}   & 56.4 & \textbf{53.1}          \\ \hline
\end{tabular}
\end{table}			

\begin{figure}[t]
\centering
\includegraphics[scale=0.37]{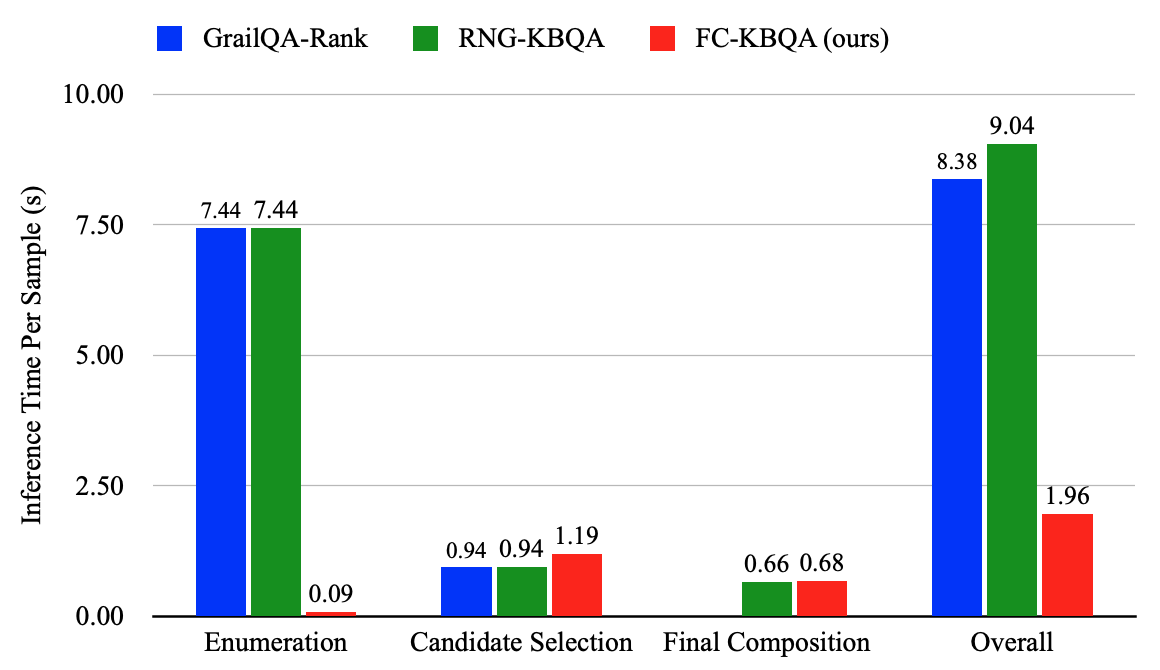}
\caption{Inference time on GrailQA. ``Overall'' denotes the total inference time of each model. Specially, GrailQA-Rank has no composition step, we record the corresponding time as zero. }     
\label{fig:inference_time}

\end{figure}

\vpara{Efficiency.}
Both RNG-KBQA and GrailQA-Rank enumerate all the logical expressions in a 2-hop KB subgraph (enumeration), so it is time-consuming for the rank model to score thousands of logical expressions for each question (candidate selection). Conversely, our \smodel just retrieves the most relevant components (candidate selection) and then enumerates the component pairs based on the filtered candidates (enumeration), which greatly reduces the inference time. Besides enumeration and candidate selection, a seq-to-seq model is used to generate the final logical expression (final composition).
In the same 24GB GPU and Intel Gold 5218 CPU, the experimental results in Figure~\ref{fig:inference_time} show that our model runs 4 times faster than baselines.

\begin{table*}[t]
\centering
\setlength\tabcolsep{10pt}
\caption{Ablation studies on GrailQA-Dev (\%).}
\label{tb:ablation-result}
\begin{tabular}{lcccccccc}
\hline
& \multicolumn{2}{c}{Overall}   & \multicolumn{2}{c}{I.I.D.}    & \multicolumn{2}{c}{Compositional} & \multicolumn{2}{c}{Zero-Shot} \\ \cline{2-9} 
& EM & F1 & EM  & F1  & EM  & F1  & EM   & F1   \\ \hline
\multicolumn{1}{l}{T5-base}  & 22.7   & 23.4   & 61.8          & 64.1          & 28.3           & 29.0           & 0.3          & 0.3 \\
\multicolumn{1}{l}{RNG-KBQA}  & 71.4   & 76.8   & 86.5          & 88.9          & 61.6            & 68.8            & 69.0          & 74.8          \\ 
\multicolumn{1}{l}{Enhanced RNG-KBQA} & 72.8          & 78.2          & 86.6          & 90.2          & 61.7            & 69.3            & 71.5          & 76.7          \\ \hline
\multicolumn{1}{l}{\model}   & \textbf{79.0} & \textbf{83.8} & \textbf{89.0} & \textbf{91.5} & \textbf{70.4}   & \textbf{77.3}   & \textbf{78.1} & \textbf{83.1} \\ 
\multicolumn{1}{r}{\quad--Knowledge}                   & 23.1          & 24.0          & 62.1          & 64.2          & 29.5            & 31.0            & 0.3          & 0.3          \\   
\multicolumn{1}{r}{\quad--Knowledge Pairs}                  & 53.6          & 55.6          & 70.2          & 72.3          & 44.0            & 46.0            & 50.3          & 52.2   
   \\ 
\multicolumn{1}{r}{\quad--Logical Skeleton}  & 78.0          & 80.8          & 85.2          & 86.8          & 68.5            & 71.9            & 79.2          & 81.8 \\ 
\multicolumn{1}{r}{\quad--Decode Constraint}  & 77.5          & 83.1          & 88.3          & 91.1          & 67.8            & 76.3            & 76.8          & 82.5 \\ \hline
\end{tabular}
\end{table*}

\subsection{Ablation Studies}
\label{sec:ablation}
GrailQA does not provide ground truth for the test set, so we conduct the ablation studies on the public Grail-Dev to investigate how the fine- and middle-grained components affect the performance.

As shown in Table~\ref{tb:ablation-result}, we develop four model variants. (1) \textbf{-Knowledge} removes all the fine-grained and middle-grained components except for the logical skeleton. (2) \textbf{-Knowledge Pairs} replaces the middle-grained components, such as  class-relation pairs and relation-relation pairs with the corresponding fine-grained candidates, such as classes and relations. (3) \textbf{-Logical Skeleton} gets rid of the logical skeleton. 
(4) \textbf{-Decode Constraint} deletes the dynamic vocabulary created with the middle-grained components. 

The results show that  removing ``knowledge'' reduces model performance by 60\% F1 score, and replacing  ``knowledge pairs'' with pure fine-grained components also reduces model performance by 28\% F1, indicating that encoding the middle-grained components can significantly improve the model's reasoning capacity. To further demonstrate that encoding such middle-grained components can also help improve other model's performance, we create Enhanced RNG-KBQA by taking the top-10 ranked results from its ranking model and formulating them into middle-grained component pairs to be injected into its encoder. The results in Table~\ref{tb:ablation-result} show that middle-grained reformulation improves the performance of RNG-KBQA.
Middle-grained component pairs, like coarse-grained logical expressions, can guarantee connectivity, but they are more compact and much shorter.
As a result, because PLMs have a maximum input length, the middle-grained formulation can inject more components and is more likely to cover the components involved in the target logical expression.

Removing ``logical skeleton'' can result in a 3.0\% F1 drop, indicating that skeleton is useful for guiding the question understanding even though it is less important than the knowledge.

Removing ``decode constraint'' in the decoder can also have an effect on model performance, but is much weaker than removing ``knowledge pairs'' in the encoder, indicating that injecting the knowledge constraints in the encoding process is more useful than in the decoding process, because the encoder learns the bidirectional context, which is better suited to natural language understanding. This is also a significant difference from the existing knowledge constrained decoding methods.

Both "Knowledge Pairs" and "Decode Constraint" are proposed for addressing the in-executability issue, which guarantee all generated logical expressions are executable. Removing either reduces the accuracy, which indicates that high executability can improve the model performance.

\subsection{Error Analysis}
We randomly select 50 error cases on GrailQA and summarize the error into three main categories: error entity (60\%), error relation and class (35\%), and error logical skeleton (40\%). We also analysis the error cases while our model fails but some baseline methods can answer successfully resolve them. A typical mistake is on logical expressions that involve KB-specific component composition. For example, in Freebase, ``coach'' is represented by the join of ``sports.sports\_team.coaches'' and ``sports.sports\_team\_coach\_tenure.coach''. Our fine-to-coarse model only predicts the  previous relation but is unable to recall ``sports.sports\_team\_coach\_tenure.coach'', while some coarse-grained methods are able to memorize such composition and provide the correct answer.

\section{Conclusion}
This paper proposes \model, a \underline{F}ine-to-\underline{C}oarse composition framework for KBQA. The core idea behind it is to solve the 
entanglement issue of mainstream coarse-grained modeling by the fine-grained modeling, and further improve the executability of logical expression by reformulating the fine-grained knowledge into middle-grained knowledge pairs.
Benefiting from this, \smodel achieves new state-of-the-art performance and efficiency on the compositional and zero-shot generalization KBQA tasks. 
This fine-to-coarse framework with middle-grained knowledge injection could be inspiring for generalization on other NLP tasks.

\section{Limitations}
\label{sec:limitation}
Although our model achieves good performance in solving the compositional and zero-shot generalization problems, there is still room for improvement on the i.i.d datasets. The fine-grained module in our framework cannot take advantage of explicit composition information when the component compositions in the testing set and training set significantly overlapp. For example, in Freebase, "Who is the coach of FC Barcelona?" is answered by the join of relation ``sports.sports\_team.coaches''
and ``sports.sports\_team\_coach\_tenure.coach''. 
Our fine-grained extractor may fail to recall ``sports.sports\_team\_coach\_tenure.coach'' and instead select ``base.american\_football.football\_coac\\-h.coach'' as the candidate  since `football coach'' is more relevant to the question than ``coach tenure'' in semantics. The only coarse-grained model, however, can directly memorize the pattern because such composition appears frequently in the training data. Therefore, compared to conventional models that completely memorize composition patterns, our model may only have minor advantages.

Another limitation is that we cannot guarantee the generalization on other KBs such as WikiData because gaps between KBs may bring negative impact. For example, relations in Freebase are often more specific (ice\_hockey.hockey\_player.hockey\_position, soccer.football\_player.position\_s), while relations in Wikidata are more general (position\_played\_on\_team). We consider it as a direction for our future work.

\section{Ethics Statement}
This work focuses on the generalization issue of knowledge base question answering, and the contribution is fully methodological. Hence, there are no direct negative social impacts of this work. For experiments, this work uses open datasets that have been widely used in previous work and are without sensitive information as we know. The authors of this work follow the ACL Code of Ethics and the application of this work have no obvious issue that may lead to the risk of ethics.

\section*{Acknowledgments}
This work is supported by National Natural Science Foundation of China (62076245, 62072460, 62172424,62276270); Beijing Natural Science Foundation (4212022).
\bibliography{anthology,custom}
\bibliographystyle{acl_natbib}

\appendix
\section{Implementation Detail}
\vpara{KB Environment.} To execute the SPARQL, we build a virtuoso database with the latest official data dump of Freebase\footnote{https://developers.google.com/freebase}.

\vpara{Pilot Study.}
To simulate the generalization problems, the training set and test set are drawn from GrailQA's training set and test set, respectively.
To build the toy train set, we choose two thousand cases with only the 1-hop logical expression from the GrailQA train set. In addition, for the compositional test set, we select the 1-hop cases from the GrailQA test set, which contains seen single relations and classes but unseen class-relation pairs beyond the train set. For the zero-shot test set, we select the 1-hop cases that involve both a class and a relation that does not appear in the toy train set. To be noted, as coarse-grained modeling involves the enumeration of logical expressions to obtain candidates, and the enumeration is nearly impossible for 2-hop logical expressions due to the large amount (greater than 2,000,000). So, we simplify the pilot study to only 1-hop questions that involve the composition of a class and a relation, which can also support comparing fine-grained and coarse-grained modeling.

For both the coarse-level and fine-level matching methods, we apply a BERT-based-uncased model. Both models are trained for 5 epochs with a batch size of 8 and a learning rate of 2e-5. To demonstrate the capacity of the models and make an objective comparison, we also employ the contractive loss with a random negative sample for both strategies.

\vpara{Extraction Model.} For both the relation extractor and class extractor, we also apply the BERT-based-uncased model. The encoder accepts the concatenation of the question $q$ and relation $r$ or the class $c$ as the input, and then a linear layer projects the output [CLS] embedding into a similarity score $s(q, r)$ or $s(q, c)$. The BERT is fine-tuned by optimizing a contrastive loss~\cite{chen2020simple},

\begin{equation}
\mathcal{L}\left({q, r_{pos}}\right) = -\text{log}\frac{e^{s(q, r_{pos})}}{e^{s(q, r_{pos})}+\sum_{r^{'} \in \{r_{neg}\}}{e^{s(q, r^{'})}}}  \nonumber
\label{extractor_loss}
\end{equation}

\noindent where $r_{pos}$ is one of the golden relations extracted from the target logical expression, and $\{r_{neg}\}$ is the set of the negative relations sampled from relation set which shares the same domain as $r_{pos}$. We sample 48 negative candidates for each sample and fine-tune BERT-base-uncased for 10 epochs with a batch size of 8 and a learning rate of 2e-5.

\vpara{Generation Model.} We initiate both of our seq-to-seq models with T5-based provided by the huggingface library~\cite{wolf-etal-2020-transformers}. For logical skeleton parsing, we fine-tune for 5 epochs with a batch size of 4 and a 4-step gradient accumulation. For the final composition model, we fine-tune for 10 epochs with a batch size of 8 and a 4-step gradient accumulation. To be noted, both the designed rules for logical skeleton parsing and vocabulary constraints in decoding process will not be used in the training process, and both training object follow the regular BART.

\section{Component Detection Models.}

\begin{table}[t]
\caption{Entity linking accuracy (\%).}
\label{tb:entity-link}
\centering
\begin{tabular}{lc}
\hline
                              & Accuracy     \\ \hline
GrailQA                       & 68.0          \\
RNG-KBQA                      & 81.6          \\ \hline
Ours                          & \textbf{87.2} \\
\quad--Relation-aware Pruning & 83.0          \\ \hline
\end{tabular}
\end{table}

\vpara{Entity Linking.}
As shown in Figure~\ref{tb:entity-link}, compared with the entity linking (EL) strategy in RNG-KBQA, our EL strategy gains 5.6\% accuracy improvement. The reasons include (1) the trie tree considers all entities' surface names, ensuring the high coverage of entity candidates, (2) the relation-aware pruning strategy can effectively remove hard negatives with similar mentions but completely different semantics. 

\begin{figure*}[th]
\centering
\includegraphics[scale=0.45]{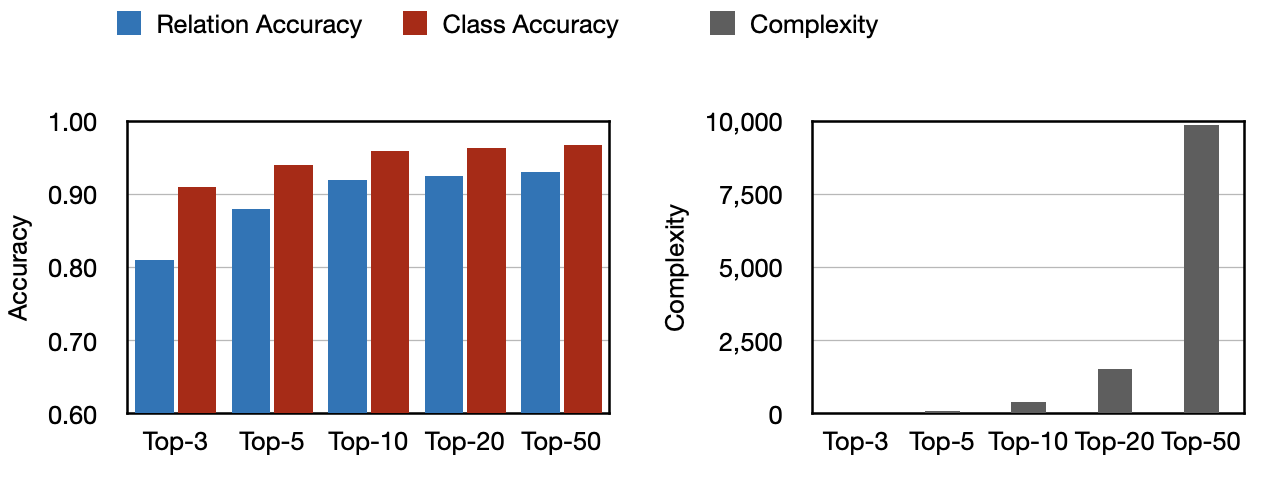}
\caption{Performance of top-$k$ relation and class extraction. Accuracy denotes the coverage of relation or class candidates. Complexity denotes the number of compositions that should be enumerated. }     \label{fig:relation}
\end{figure*}

\vpara{Relation and Class Extraction.}
Figure~\ref{fig:relation} depicts the effects of varying different sizes ($k$) of relations and classes. With the increase of $k$, the relation or class coverage represented by accuracy begins to grow slowly and attends to be stable when $k$ is around 10. Meanwhile, the complexity of composition enumeration grows exponentially with $k$. Thus, to balance efficiency and performance, we choose top-10 relations and top-10 classes.

\vpara{Logical Skeleton Parsing.}
 Table~\ref{tb:logical-form} displays the effectiveness of logical skeleton parsing techniques for various beam searches. ``Raw Question'' refers to directly parsing the raw question into the logical skeleton, while ``+Mask'' refers to parsing using our entity mask strategy. 
 For both the strategies, in addition to the top-1, top-2, and top-3 beam search results, we also report the results of Top-2(R) which add ``<rel><rel>'' as the top-2 candidate if ``<rel>'' is the top-1 prediction, vice versa. We can see that our designed entity mask strategy and rule-based beam search can contribute to the logical skeleton parsing. The rules significantly improve the performance as 1-hop relation and 2-hop relations are quite mix up in KBs. For example, the semantic-grained one-hop relation ``program producer'' could be represented by a 1-hop relation (``tv.tv\_producer.programs\_produced'' in domain TV) or a 2-hop relations (``broadcast.content.producer'' and ``radio.radio\_subject.p- rograms\_with\_this\_subject'' in domain radio). 

\section{Running Example}
We here give a running example of our framework for better understanding. As shown in Figure~\ref{fig:over-view}, given the question ``the terminuses of Antonio belongs to what railway?'', we first propose fine-grained component detection. We retrieve candidate classes ``railway'', ``railway\_terminus'', ``railway\_type'', ... and candidate relations ``railway.terminuses'', ``railway.branches\_to'', ``transit\_line.terminuses'',..., and candidate entities ``Antonio'' which is a football player,``Antonio'' which is a city ,..., and logical skeleton candidates. Then, we apply the middle-grained constrain, for example, for class-relation pairs, ``railway'' is connected to ``railway.terminuses'' in KB but not connected to ``railway.branches\_to''; for relation-relation pairs,``railway.terminuses'' shares matched domain and range with ``railway.branches\_to'' but not share with ``transit\_line.terminuses''; for entities, the football player ``Antonio'' does not match any candidate relations and will be pruned.  Finally, we put question, all connected class-relation pairs, all connected relation-relation pairs, all entities that have not been pruned and logical skeleton candidates into the composition model and generate logical expression.
\section{Case Study}
\begin{figure*}[t]
    \includegraphics[scale=0.7]{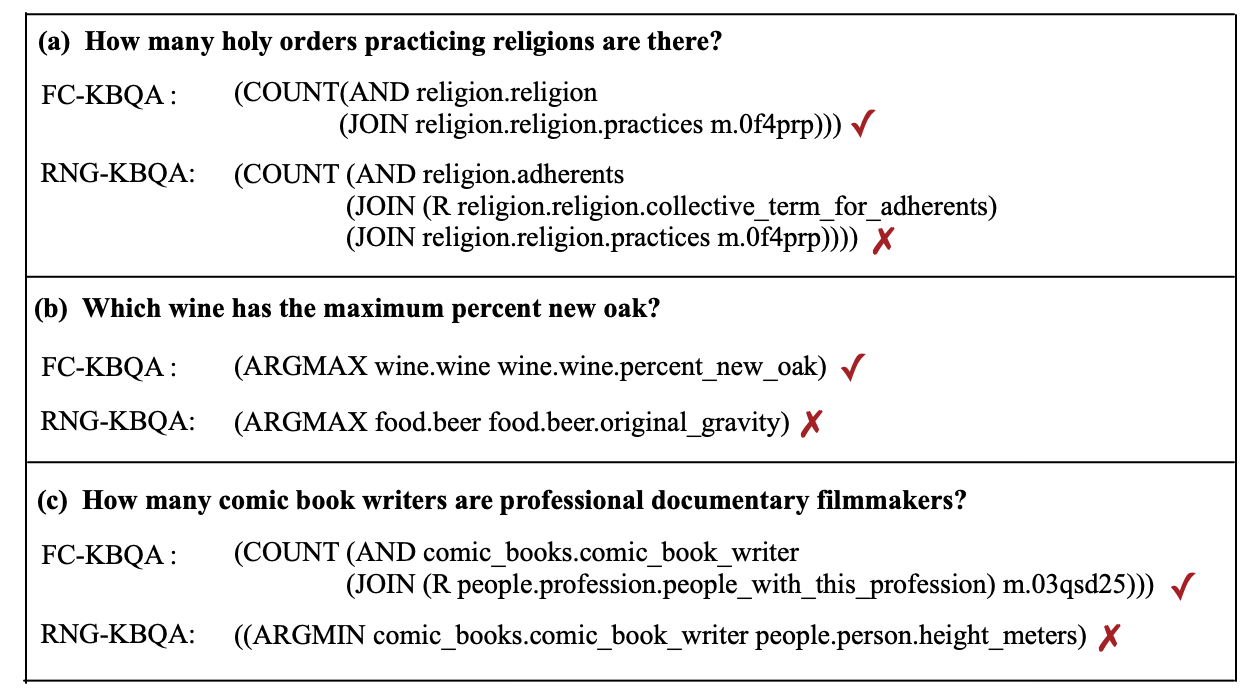}
    \caption{Case Study on GrailQA.}
    \centering
    \label{fig:case-study}
\end{figure*}

\begin{table}[t]
\caption{Logical skeleton parsing accuracy (\%).}
\label{tb:logical-form}
\centering
\setlength{\tabcolsep}{1mm}{
\begin{tabular}{lcccc}
\hline
& \multicolumn{1}{l}{Top-1} & \multicolumn{1}{l}{Top-2} & \multicolumn{1}{l}{Top-3} & \multicolumn{1}{l}{Top-2(R)} \\ \hline
Raw Question   & 83.2    & 86.1     & 86.7  & 94.0 \\
\multicolumn{1}{r}{+Mask} & 85.5 & 87.4 & 88.6 & \textbf{95.3} \\
\hline
\end{tabular}}
\end{table}
Figure~\ref{fig:case-study} shows some cases that our \smodel and RNG-KBQA predicted. 
Example(a) shows a simple one-hop case, but RNG-KBQA tends to generate a more complex logical expression because it frequently occurs in the training set. 
With sample cases where the surface name of the gold relation has a clear overlap with the question, Example(b) demonstrates how the composition of each component causes RNG-KBQA to fail. 
As seen in example(c), the entanglement of knowledge and logical skeleton causes RNG-KBQA to predict some straightforward logical operators like "COUNT" incorrectly. These restrictions can be overcome by our proposed \model.

\end{document}